\documentclass[journal]{IEEEtran}
\usepackage{cite}
\usepackage{amsmath,amssymb,amsfonts}
\usepackage{epstopdf}
\usepackage{graphicx}
\usepackage{textcomp}
\usepackage{hyperref}
\usepackage{floatrow}
\usepackage{algorithm}
\usepackage{algpseudocode}
\usepackage{float}
\usepackage{multirow} 
\usepackage{subfigure}
\usepackage{threeparttable}
\usepackage{longtable,lscape}
\usepackage[table,xcdraw]{xcolor}
\usepackage[justification=centering]{caption}
\usepackage{footnote}
\usepackage{booktabs}
\newfloatcommand{capbtabbox}{table}[][\FBwidth]
\usepackage{url}
\usepackage{multicol}
\begin{document}

\title{FedPDC:Federated Learning for Public Dataset Correction}

\author{\IEEEauthorblockN{Yuquan Zhang,
Yongquan Zhang\thanks{* Yongquan Zhang is the corresponding author.}\IEEEauthorrefmark{1}
}}

\markboth{}%
{}

\maketitle

\begin{abstract}
As people pay more and more attention to privacy protection, Federated Learning (FL), as a promising distributed machine learning paradigm, is receiving more and more attention. However, due to the biased distribution of data on devices in real life, federated learning has lower classification accuracy than traditional machine learning in Non-IID scenarios. Although there are many optimization algorithms, the local model aggregation in the parameter server is still relatively traditional. In this paper, a new algorithm FedPDC is proposed to optimize the aggregation mode of local models and the loss function of local training by using the shared data sets in some industries. In many benchmark experiments, FedPDC can effectively improve the accuracy of the global model in the case of extremely unbalanced data distribution, while ensuring the privacy of the client data. At the same time, the accuracy improvement of FedPDC does not bring additional communication costs.
\end{abstract}


\IEEEpeerreviewmaketitle
\setlength{\parskip}{0.8em}
\section{Introduction}\label{a}

Deep learning has been widely used in many tasks involving images, video, voice and so on and has achieved great success. However, deep learning usually requires a lot of data to train the model, so as to obtain a better parameter model. At the same time, training at the same terminal also requires a lot of time and computing resources. Some small companies do not have so much data and computing power. Besides, with the development of big data, it has become a worldwide trend to pay attention to data privacy and security. Every time the public data is leaked, the media and the public will pay great attention to it. This is obviously a severe blow to our traditional machine learning, because the previous traditional machine learning relied on massive data to train a good model, but now there is a barrier between the data, which is the origin of the phenomenon of data islands.

How to design a machine learning framework under the premise of meeting the requirements of data privacy, security and supervision, so that artificial intelligence systems can use their own data more efficiently and more accurately, is an important topic in the current development of artificial intelligence. We propose to shift the research focus to how to solve the problem of data islands. We propose a feasible solution for privacy protection and data security both, called federated learning(FL)\cite{b1}.

Federated learning is a kind of distributed machine learning, which is based on parallel computing. It differs from distributed machine learning in that it does not exchange data and gradients in the communication process of federal learning, which ensures participants'privacy. The goal of Federated learning is to solve the problems of data collaboration and privacy protection. The distribution of Federated learning's data is not independent and identically distributed, because there are differences between users, and the amount of data may not be an order of magnitude. So it does not conform to the probability distribution of the independent identical distribution.

Federated learning is a distributed learning paradigm with two key challenges different from traditional distributed optimization\cite{b2}: (1) significant variability (system heterogeneity) in system characteristics on each device in the network, and (2) non uniformly distributed data (statistical heterogeneity) across the network.

\begin{figure*}[ht]
    \centering
    \includegraphics[scale=0.5]{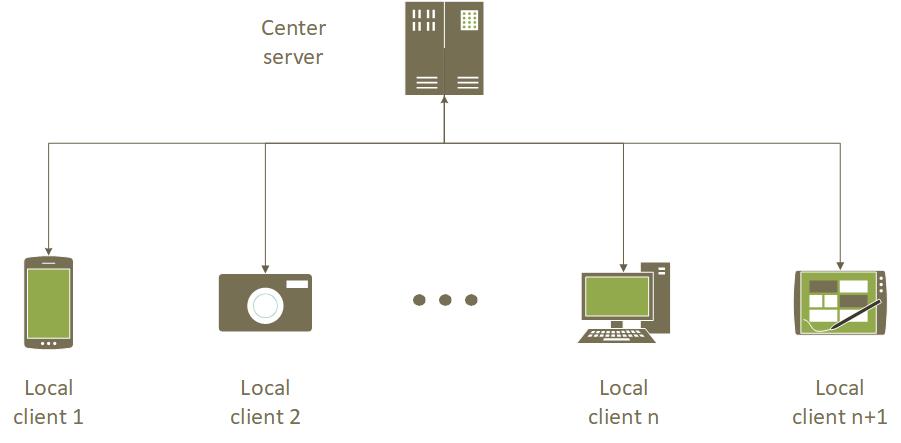}
    \caption{The framework of classical Federated learning in our real life.}\label{fig1}
\end{figure*}

In order to deal with heterogeneity and address high communication costs, optimization methods that allow local updating and low participation are a popular method of Federated learning. Such as FedAvg\cite{b1}, FedProx\cite{b2} and so on. These methods effectively protect the privacy of participants, and under ideal conditions, the accuracy rate can be almost the same as that of centralized training. But sometimes the reality is that the data is extremely unbalanced, just like the chest X-ray of patients with COVID-19 infection. As we know COVID-19 is highly infectious and the incidence of infection is regional. As a result, hospitals in some regions almost have no lung X-ray films of infected persons as a reference, and these hospitals are unable to obtain lung X-ray films of infected persons in other hospitals based on the requirements of privacy protection, which makes it difficult to judge whether the patient is infected with COVID-19. However, these hospitals may receive some patients' X-ray films from the World Health Organization or other public health departments. Therefore, our method can improve the prediction accuracy of the global model by using the accuracy of the local model on a small number of images with balanced data distribution.\\

\noindent Our contributions are primarily summarized as follows:
\begin{itemize}
    \item [*] We propose FedPDC based on the model correction of the verification accuracy of the central server. The classification accuracy of each local model on the server is used for weighted average to obtain a new round of global model. And The accuracy is added to the training as the regular term of the loss function in the next round.
    \item [*]Our approach is robust to typical challenges of federated learning including data heterogeneity and low client participation rate, and is applicable to various federated learning frameworks.
    \item [*]We have proved through a large number of experiments that, compared with the most advanced federated learning algorithm, our architecture regularization technology consistently improves the accuracy and convergence speed.
\end{itemize}

\section{Related work}\label{e}
In federated learning, there are two main research directions: horizontal federated learning and vertical federated learning\cite{b4}. Among them, horizontal federal learning faces several challenges.

The model transmission between the client and the parameter server has a huge communication consumption.Mcmahan H B ,  Moore E ,  Ramage D , et al. proposed FedAvg\cite{b1}, the number of local training epochs is increased to reduce communication consumption. So compared with FedSGD, FedAvg greatly reduces communication consumption while maintaining almost the same accuracy. And the convergence of FedAvg has been proved\cite{b17}, which formally initiated the concept of federated learning. Zhu L ,  Lin H ,  Lu Y , et al. proposed DGA\cite{b5}, The aggregation of the delay gradient is used to alleviate the long time of the aggregation global model caused by the high delay of some clients. Reisizadeh A ,  Mokhtari A ,  Hassani H , et al. proposed FedPAQ\cite{b7}, the client in the network is allowed to perform local training before synchronizing with the central server, and only the updates of the active client are sent back to the central server, and only the quantized version of the local information is sent back. These algorithms have effectively saved the communication loss, and even some can avoid the training failure caused by some clients offline.

The uneven distribution of local data per client is another challenge\cite{b10}. In order to improve the classification accuracy of the global model on Non-IID data, there are three main categories of methods: (1) The update of the local model is corrected based on the update direction of the global model. (2) Grouping clients based on the similarity of local models. (3) Knowledge distillation. 

The updating direction correction based on the global model needs to add its similarity to the loss function in the local model training process. Li T ,  Sahu A K ,  Zaheer M , et al. proposed FedProx\cite{b2}. In the local model training, the near term is added to regularize the local loss function. Li Q ,  He B ,  Song D proposed a new framework named MOON\cite{b3}, it corrects local updates by maximizing the consistency between the current local model learning representation and the global model learning representation. Karimireddy S P ,  Kale S ,  Mohri M , et al. proposed a random control averaging scheme called scaffold \cite{b6}, which can solve the problem of "client drift" caused by data heterogeneity. However, the introduction of global control variables in scaffold will lead to higher communication overhead, which is twice that of fedavg.

FedEntropy\cite{b8} is a typical example of solving data heterogeneous problems by grouping. Grouping based on the maximum entropy variation and purposefully selecting clients. Duan M ,  Liu D ,  Chen X , et al. proposed Astraea\cite{b9}, the local models are grouped based on the KL divergence of the data distribution of the local models to rearrange the training of the local models.

knowledge distillation originally proposed by Bucila, Caruana, and Niculescu-Mizil (2006) \cite{b20} and popularized by Hinton, Vinyals, and Dean(2015)\cite{b21}. Knowledge distillation compresses the knowledge of a large, computationally expensive model into a computationally efficient neural network. The idea of knowledge rectification is to train small models and students on a transfer set of soft targets provided by large models and teachers. Li et al. \cite{b19} proposed FedMD to tackle the heterogeneous models through knowledge distillation. Ganta D P et al. proposed a new knowledge distillation scheme and introduced the intermediate model of teaching assistant to improve the accuracy loss caused by the compression model\cite{b18}.

In addition, federal learning also has personalized methods to solve some extremely unbalanced data problems\cite{b24, b25, b26, b27, b28, b29}. M G. Arivazhagan et al. proposed the algorithm of adding personalization layer in federated learning\cite{b22}. L. Collins et al. regard the data heterogeneous federation learning problem as n parallel learning tasks that may have some common structure, and learn and use this common representation to improve the quality of each client model\cite{b23}.

Although the above FL methods are promising, they do not take into account the difference between the data distribution characteristics of devices and the overall data distribution characteristics, which can be used to further improve the classification performance in Non-IID scenarios. As far as we know, our work is the first attempt to apply a typical small amount of public data sets to the server to improve the accuracy of the global model. Since our method makes full use of the difference between the distribution of device data in Non-IID scenarios and that of IID data, it can not only improve the overall FL classification performance, but also not cause additional communication loss.

\section{Preliminaries}\label{b}

\subsection{Federated learning}\label{b.1}
Federated learning is a distributed machine learning technology that allows multiple parties to build models according to specified algorithms through local training sets. For example, In the medical field, we usually have stricter privacy protection requirements than in other fields. Suppose that in the prediction and prevention of the same disease, each hospital can not leak its users' private data to each other, such as diagnosis and treatment records and X-rays. But local data from each hospital can be unbalanced, and the amount of data is too small to train a model that works well. At this point we use federated learning, where we treat each hospital as a client, and the data between each client is not communicative to each other. We set up a parameter server to receive, aggregate, and send each round of the model. This parameter server also does not receive any patient data, thus ensuring privacy. Federated learning aggregates locally trained models from each hospital to get a global model that predicts better. 

Specifically, the federated learning process is that after the federated learning participants have trained the local data, the parameters and the models obtained from the training are uploaded to the server, and The server receives the models and aggregates it, distributing the aggregated model to each client. This is the process of a communication round of federated learning. The final convergence model is obtained after several communication rounds.

Therefore, we can simplify the main optimization of a federated learning as follows:
considering $N$ clients $C_{1},...,C_{N}$ where each client $C_{i}$ have a local dataset $\mathcal{D}_{i}$, federated learning is to learn a global model $w$ over the all dataset $\mathcal{D}\triangleq\cup_{i\in[N]}\mathcal{D}_{i}$ with the parameter server $P$ which have no local data $P\cap \mathcal{D}=\varnothing$.The optimization goal of federated learning can be defined as
\begin{equation}
 	\label{eq1}
 	\mathop{\arg\min}\limits_{w}\ \ \mathcal{L}(w)=\sum_{i=1}^N\dfrac{|\mathcal{D}_{i}|}{|\mathcal{D}|}L_{i}(w),
 \end{equation}
where $|\cdot|$ denotes the cardinality of sets and $L_{i}(w)=\mathbb{E}_{(x,y)\sim\mathcal{D}_{i}}[\ell_{i}(w;(x,y))]$ is the expectation for the cross-entropy loss function $\ell_{i}$ of the client $C_{i}$ in classification problems. Here, $(x,y)$ denotes the input data $x$ and its corresponding label $y$\cite{b1}.

\subsection{Federated Averaging (FedAvg)}\label{b.1}

The Federated Average is one of the most classic Federated Learning algorithms, and it improved on top of FedSGD.FedSGD in each communication round, each client performs only one epoch local training. But since federated learning is in most cases subject to the latency of communication and the bandwidth of communication rather than the time required for local training. So FedSGD's training efficiency is not high. FedAvg's proposal alleviates this problem by letting the client train several epochs locally in each communication round and upload the last local model to the parameter server for aggregation.At the same time, in order to avoid all clients uploading the model when some clients are down or cause network traffic congestion, the rest of the clients can not train the next communication round, FedAvg set a hyperparameter $\tau\in \left ( 0,1 \right ] $, Only $N\times \tau$ (when $\tau$ = 1, FedAvg equivalent to FedSGD)clients per communication round upload their local model of that last epoch to the parameter server. 
\begin{figure}[ht]
    \centering
    \includegraphics[scale=0.3]{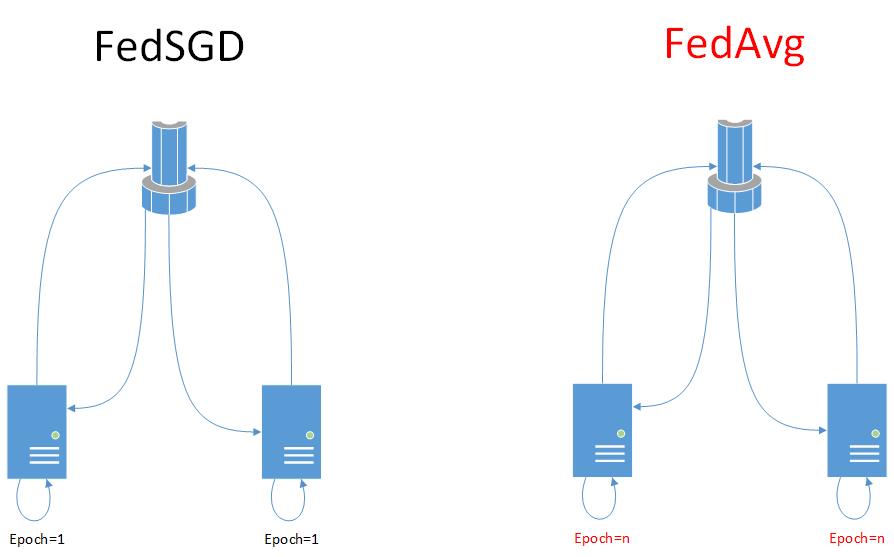}
    \caption{The left on is FedSGD and the right one is FedAvg.}\label{fig2}
\end{figure}
In each communication round $\emph{t}$, the parameter server randomly selects $S_{r}=N\times \tau$ clients for this round of local training. The parameter server sends the global model of this communication round $w ^{t}$ to these clients, and these clients locally use optimization algorithms such as SGD to train the model, and the iterative process is
 \begin{equation}
 		\label{eq2}
w_{i}^{t+1}(m)=w_{i}^{t+1}(m-1)-\eta\nabla L_{i}(w_{i}^{t+1}(m-1))
\end{equation}
where $w_{i}^{t+1}(m)$ represents the model of the $\emph{i}$th client in the $\emph{m}$th local epoch in the $\emph{t+1}$th communication round, $L_{i}(\cdot)$ is the cross-entropy loss function or other loss function, and $\eta$ is the learning rate.

After $M$ times local updates, each client sents $w_{i}^{r}(\tau)$ to the server which updates the global model $w^{t+1}$ as
 \begin{equation}
 		\label{eq3}
w^{t+1}=\sum_{i\in S_{r} }\frac{|\mathcal{D}_{i}|}{|\mathcal{D}_{S_{r}}|} w_{i}^{t+1}(M),
\end{equation}
with $\mathcal{D}_{S_{r}}\triangleq\cup_{i\in[S_{r}]}\mathcal{D}_{i}$.

\renewcommand{\algorithmicrequire}{\textbf{Input:}}  
\renewcommand{\algorithmicensure}{\textbf{Output:}} 
\begin{algorithm}[h]  
  \caption{\large FedAvg}  
  \label{fedavg}
  \begin{algorithmic}
    \Require{The $\emph{K}$ clients are indexed by $k$; $b$ is the local minibatch size, $T$ is the number of local epochs, and $\eta$ is the learning rate.}\label{alg:fedavg}
    \Ensure  
   The final model $w^{T}$ 
   \State \textbf{server executes:}
    \State initialize $w_0$
    \For{each round $t = 1, 2, \dots$}
        \State $m \leftarrow \max(\tau\cdot K, 1)$
        \State $S_t \leftarrow$ (random set of $m$ clients)
        \For{each client $k \in S_t$ \textbf{in parallel}}
            \State $w_{t+1}^k \leftarrow \text{\textbf{client executes}}(k, w_t)$
        \EndFor
        \State $w_{t+1} \leftarrow \sum_{k=1}^{K} \frac{n_{k}}{n} w_{t+1}^{k}$
    \EndFor
    \State \textbf{client executes:}
    \State $\mathcal{B} \leftarrow$ (split $P_{k}$ into batches of size $B$)
        \For{each local epoch $i$ from $1$ to $M$}
            \For{batch $b \in \mathcal{B}$}
                \State $w \leftarrow w - \eta \nabla \ell(w ; b)$
            \EndFor
        \EndFor
    \State return $w$ to server
  \end{algorithmic}  
\end{algorithm}

\section{Method}\label{c}

As mentioned above, in the federated learning of real problems, most of the data owned by the client has high heterogeneity and uneven distribution, which is also the main obstacle affecting the model accuracy in the federated learning. However, there are many common shared data in many industries and fields, and these data are often distributed evenly and have low heterogeneity.Therefore, in our algorithm, we make full use of these unbiased data to correct the local model, so that its update direction is closer to the global model and has higher accuracy.

Our algorithm uses the public data set in two aspects, the aggregation strategy optimization at the parameter server side and the loss function optimization during local training.

\subsection{Global precision aggregation}\label{c.1}

In most classical federated learning algorithms, the aggregation algorithm of parameter server usually adopts the strategy of weighted average according to the data volume of each client. But these simple weighted averages also have obvious disadvantages, because most of the data owned by the client in the real task is unbalanced. If we simply use these unbalanced data to carry out simple weighted averaging, the model will only over learn those features that appear most frequently, and those features that appear less frequently will be further diluted. Using this biased global model to predict real problems will lead to a large deviation in the prediction results.

In our algorithm, in each communication round I, we use the existing unbiased data set to verify in the parameter server to correct the local model and obtain better robustness. Like Fedavg and other classical algorithms, every communication round $t$, the parameter server sends the global model $w^{t}$ of the round to each client $i$. Each client receives the global model $w^{t}$ and then conducts local training for $M$ epochs. After $M$ epochs of training, the local model of the $i$th client is obtained $w_{i}^{t}$. The selected client sends the local model to the parameter server. There is an unbiased small data set $S_{s}$ in the parameter server. Each local model was used to predict the data set, and the accuracy was $p_{i}^{t}$. Then our aggregation strategy is:

\begin{equation}
 		\label{eq4}
w^{t+1} \leftarrow \sum_{i=1}^{N} \frac{\left|p_{i}^{t}\right|}{\left|\sum_{i=1}^{\mathcal{N}} p_{i}^{t}\right|} w_{i}^{t}
\end{equation}
where $p_{i}^{t}$ is in the $t$th communication round, the local model of the $i$th client predicts the accuracy of the results of $S_{s}$. 

Because the model is trained in the local training data set, it can only fit the local data well, but cannot fit the global data well. Therefore, when there are few or no categories in the local dataset, the effect of model prediction will be poor. The aggregation method used by our algorithm is to weighted average the model through the verification accuracy on the public unbiased data set. This aggregation method effectively avoids the global model deviation caused by simple weighted average. Because the correction term used is the prediction accuracy from unbiased data sets, we can give more weight to more representative local models in aggregation. The local model trained by the clients with large amount of local data but single feature distribution will not occupy a large proportion as in the classical aggregation algorithm. Experiments show that the aggregation method can effectively improve the prediction accuracy of the global model.

\renewcommand{\algorithmicrequire}{\textbf{Input:}}  
\renewcommand{\algorithmicensure}{\textbf{Output:}} 
\begin{algorithm}[h]  
  \caption{\large FedPDC}  
  \label{alg:Framwork}  
  \begin{algorithmic}  
    \Require  
    number of communication rounds T, number of parties N, number of local epochs E, learning rate $\eta$ , hyper-parameter $\mu$  ,server-train dataset S(s)
    \Ensure  
  The final model $w^{T}$ \\
  \textbf{server executes:}
      \For{t = 0, 1, ..., T - 1}
      \State test $w_{i}^{t}$ in S(s) $\longrightarrow$ local accuracy $p_{i}^{t}$
      \For{i = 1, 2, ..., N \textbf{in parallel}} 
      \State send the global model $w^{t}$ , global accuracy $p_{t}$ to $N_{i}$ 
      \State $w_{i}^{t}$ ← \textbf{PartyLocalTraining}(i, $w^{t}$, $p_{t}$)
      \EndFor
      \State $w^{t+1} \leftarrow \sum_{k=1}^{N} \frac{\left|p_{t}^{i}\right|}{\left|\sum_{i=1}^{\mathcal{N}} p_{t}^{i}\right|} w_{k}^{t}$
      \EndFor
      \State \textbf{return $w^{T}$}
      \State \textbf{PartyLocalTraining(i, $w^{t}$, $p_{t}$):}
      \State $w_{i}^{t} \leftarrow w^{t}$
      \For{epoch i = 0, 1, ..., E}
      \For{each batch \textbf{b} = \{x, y\} of $D^{i}$}
      \State $\ell_{sup} \leftarrow \text { CrossEntropyLoss }\left(F_{w_{i}^{t}}(x), y\right)$
      \State $\ell_{con} \leftarrow p_{t}$
      \State $\ell \leftarrow \ell_{\text {sup }}+\mu \ell_{\text {con }}$
      \State $w_{i}^{t} \leftarrow w_{i}^{t}-\eta \nabla \ell$
      \EndFor
      \EndFor
      \State \textbf{return $w_{i}^{t}$ to server}
  \end{algorithmic}  
\end{algorithm}

\subsection{Global precision loss}\label{c.2}

The classical federated learning algorithm fedavg mostly uses cross entropy loss when training image local models. Other algorithms add some penalty terms to correct the local model based on the cross entropy loss. Our algorithm also proposes a new loss based on the cross entropy loss.

As mentioned above, we bring the local model $w_{i}^{t}$ into the shared data set $S_{s}$ in the parameter server to predict and obtain the accuracy $p_{i}^{t}$. $p_{i}^{t}$ can be regarded as a penalty term to measure the deviation between the local model and the expected good prediction model. At the same time, it can also represent the difference between local data and IID data, so it can be used as the penalty term of the loss function in local model training to make the local model suitable for the current local non independent identically distributed data approach the results of training under IID data during training. In our algorithm, these prediction accuracy and the global model of this round are sent to the next round of customers. After each communication round, it is necessary to reselect the client participating in this round of training, so sometimes a problem arises that we do not select the client selected in the previous round in this round. If the client selected in the next round is not selected in the previous round, the sending accuracy $p_{i}^{t}=1$. 

After that, each client has two loss functions. One is the traditional cross entropy loss $\ell_{sup}$ , the other is the local accuracy loss $\ell_{con}$.

\begin{equation}
 		\label{eq5}
\ell_{sup} \leftarrow \text { CrossEntropyLoss }\left(F_{w_{i}^{t}}(x), y\right)
\end{equation}

\begin{equation}
 		\label{eq6}
\ell_{con} \leftarrow \lambda (1-p_{i}^{t})
\end{equation}
$\ell_{con}$ will become smaller with the improvement of prediction accuracy to achieve the purpose of correcting the local model to be closer to the global model. At the same time, the accuracy of the global model in predicting those rare classes has been significantly improved. We combine the two loss functions to get 

\begin{equation}
 		\label{eq7}
\ell \leftarrow \ell_{\text {sup }}+\ell_{\text {con }}
\end{equation}

\begin{equation}
 		\label{eq8}
\ell =\text { CrossEntropyLoss }\left(F_{w_{i}^{t}}(x), y\right)+\lambda (1-p_{i}^{t})
\end{equation}
where $\lambda$ is the sensitivity parameter used to adjust the degree of correction in the loss function to a smaller number of classes in the local model. We will further explore the adaptive sensitivity parameter in section 4 of this paper. 

~\\
The specific algorithm of FedPDC is as algorithm2.

\section{Convergence analysis}\label{d}

Because FedPDC and FedAvg are architecturally nearly identical, they are also random algorithms, that is, only a few clients are selected to participate in the global aggregation in each communication wheel, and the update direction of the clients participating in the aggregation may be inaccurate. This results in FedPDC not using the same stable steps as non-random methods in the training process, but using a step-by-step reduction to converge to a stable point. So we need to quantify the difference in the local objective function between the clients.

\subsection{Local model differences}\label{d.1}

In real-world tasks, each client often has data that is not independently and equally distributed, resulting in large differences between local models. For this phenomenon, variable B is introduced to measure the similarity between local models.
\\
\\
\textbf{Definition 1} (B-local model differences) \textbf{.} The local model $w_{k}$  are B-locally dissimilar, so we define $B(w)=\frac{P}{p_{k}}$. Because the model is obtained according to the gradient descent made by the loss function, it can be further obtained $\mathbb{E}_{k}[\parallel \bigtriangledown F_{k}(w) \parallel]\le\parallel \bigtriangledown f(w) \parallel B$.
\\
\\
Here $p_{k}$ represents the server verification accuracy of the local model of the $k$th client after the communication round. $\mathit{P}$ represents the verification accuracy of the current round of global model on the server. As a sanity check, when all the local model are the same, we have $B(w) = 1$ for all $w$. However, in the federated setting, the data distributions are often heterogeneous and $B > 1$ due to sampling discrepancies even if the samples are assumed to be IID. The more unbalanced the data distribution in the client, the lower the precision of the local model's IID public dataset on the server, while the global model is relatively less affected. Thus, $B(w) \ge  1$ and the larger the value of $B(w)$, the larger is the dissimilarity among the local models.

Based on definition 1, we make a assumption that the local model is different, which will be used in the convergence analysis later. We can easily know that the verification accuracy $P\in (0,1)$ and $p_{k}\in (0,1)$. So we have made the assumption that the model dissimilarity is bounded.
\\
\\
\textbf{Assumption 1} (Bounded dissimilarity) \textbf{.} For some $\epsilon > 0$, there exists a $B_{\epsilon}$ such that for all the points $w \in \mathcal{S}_{\epsilon}^{c}$, $B(w) \leqslant B_{\epsilon}$.
\\
\\
Although samples are not IID in real-world federated learning problems, they are still sampled from incompletely unrelated distributions. Therefore, it is reasonable to assume that the differences between local models are still limited throughout the training process. 

\subsection{Non-convex FedPDC convergence analysis}\label{d.2}

Using the bounded dissimilarity assumption of the local model (Assumption 1), we now analyze the expected decrease in the objective function when performing a FedPDC step. Two other general assumptions is required before the convergence analysis.
\\
\\
\textbf{Assumption 2} (The objective function $F_{k}$ is non convex) \textbf{.} 
\\
$\exists  w \in S, F_{k}(w) \le  F_{k}(w_{0})+\langle\nabla F_{k}(w), w_{0}-w\rangle$. In the $k$th client, for all $w$, there are always $w$ and $w_{0}$ to make the above inequality hold.
\\
\\
\textbf{Assumption 3} ($L$-Lipschitz smooth) \textbf{.} Assume the functions $F_{k}$ are L-Lipschitz smooth, and there exists $L_{-}>0$, such that $\nabla^{2} F_{k} \succeq-L_{-} \mathbf{I}$, with $\bar{\mu}:=\mu-L_{-}>0$.
\\
\\
According to the above Assumptions 1,2,3, Theorem 1 (Non-convex FedPDC convergence) can be deduced.
\\
\\
\textbf{Theorem 1} (Non-convex FedPDC convergence) \textbf{.}The local model are B-dissimilar, and $B\left(w^{t}\right) \leqslant B$. It can be concluded that
\\
\\
\begin{large}
$\lambda = \frac{1}{\mu}-\frac{L B}{\bar{\mu} \mu}-\frac{L B^{2}}{2 \bar{\mu}^{2}}-\frac{2 L B^{2}}{K\bar{\mu}^{2}}+\left(1+\frac{2 L B}{\bar{\mu}}\right) \frac{\sqrt{2} B}{\bar{\mu}\sqrt{K}}>0$
\end{large}
\\
\\
Then, during iteration t of Algorithm 2, our global goal is expected to decline:
\\
\\
\begin{large}
$\mathbb{E}_{S_{t}}\left[f\left(w^{t+1}\right)\right] \leqslant f\left(w^{t}\right)-\lambda \left\|\nabla f\left(w^{t}\right)\right\|^{2}$
\end{large}
\\
\\
Where $S_{t}$ is the set of $K$ devices selected during iteration $t$.

Detailed steps are documented in Appendix A. The key step includes applying our bounded dissimilarity assumption to each subproblem, and allowing only $K$ devices to be active in each round. The last step specifically introduces $\mathbb{E}_{S_{t}}$, an expectation on equipment selection in round $t$.

\section{Experiment}\label{e}

\subsection{Hardware and environment configuration}\label{e.1}

In this experiment, we set up 10 clients and one parameter server. In order to save the training and communication time, we adopted the process of simulating the training of 10 clients on a server. The specific configuration of the server is as follows:
\\
\\
CPU: Intel Xeon W-2235, 3.80GHz\\
GPU: NVIDIA GeForce RTX 3080, 10GB\\
RAM: DDR4 2666MHz 32GB $\times$ 2
\\
\\
We use PyTorch \cite{b3} to implement FedPDC and the other baselines.

\subsection{Experimental setup}\label{e.2}

\begin{figure}[ht]
    \centering
    \includegraphics[scale=0.3]{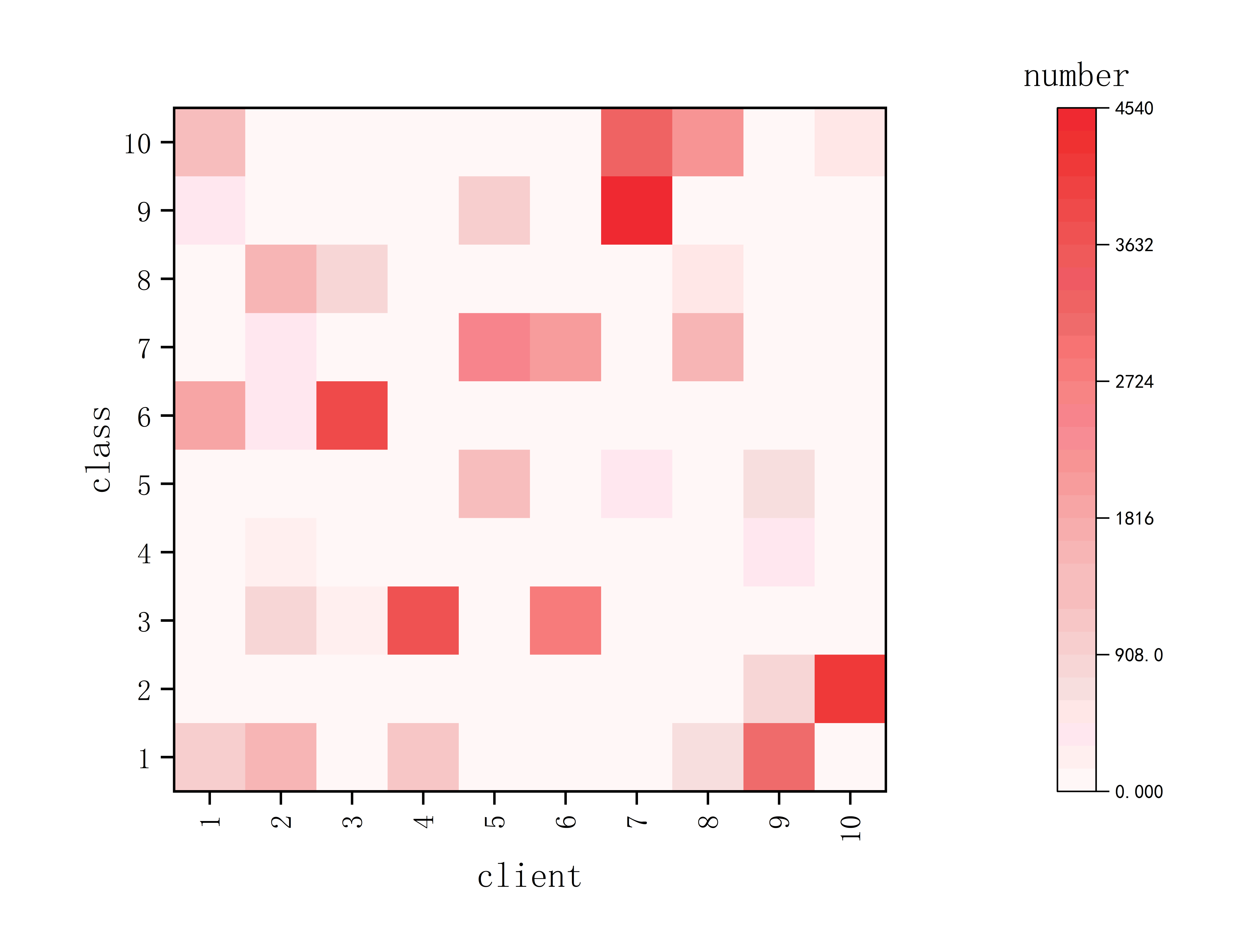}
    \caption{The number of classes in each client of CIFAR-10 dataset when Dirichlet distribution $\beta$=0.1.}\label{fig6}
\end{figure}

\begin{figure}[ht]
    \centering
    \includegraphics[scale=0.3]{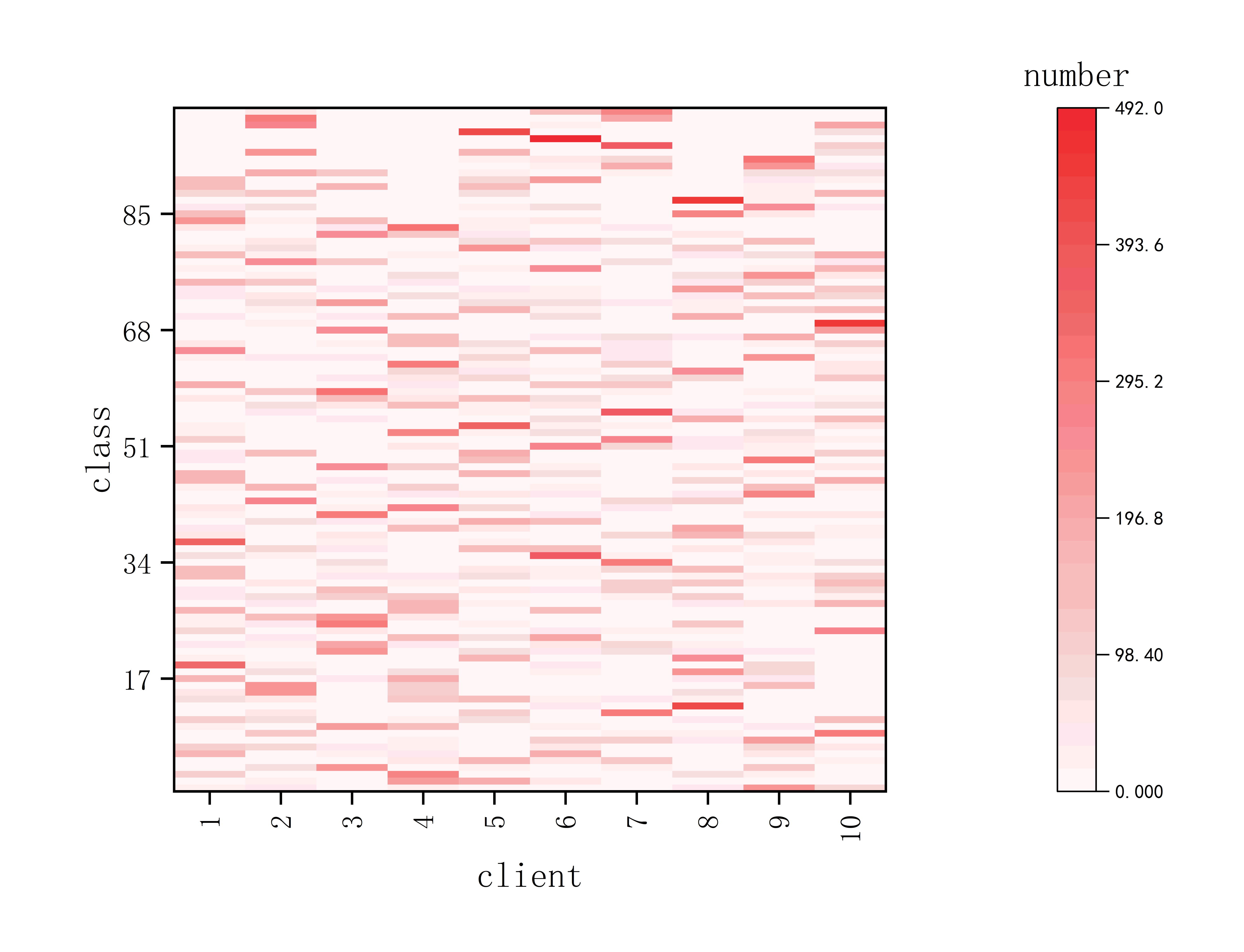}
    \caption{The number of classes in each client of CIFAR-100 dataset when Dirichlet distribution $\beta$=0.3.}\label{fig7}
\end{figure}

We compare FedPDC with three state-of-the-art approaches including (1) FedAvg \cite{b1}, (2) FedProx \cite{b2}, and(3) MOON \cite{b3}. We also compare it to the baseline, Local-Training, means that all clients do not do federated learning, but only train their own local data. We conducted experiments in three classic datasets, CIFAR-10, CIFAR-100 and Tiny-Imagenet\cite{b11}(100,000 images with 200 classes). As in the previous study\cite{b12,b13}, we use the Dirichlet distribution to generate private Non-IID data partitions for each client. Specifically, our sample $p_{k} \sim \operatorname{Dir}_{N}(\beta)$, where $p_{k}$ represents the occurrence probability of k-class samples, $\operatorname{Dir}(\beta)$ is the Dirichlet distribution with a concentration parameter $\beta$ (0.5 by default). In Fig \ref{fig6} and Fig \ref{fig7}, we can see the local data distribution of CIFAR-10 and CIFAR-100 when $\beta$ = 0.1 and 0.3 respectively. As for Tiny-Imagenet, because the number of class in this dataset is too high, there are 200 classes. In practice, we often do not encounter the situation of $\beta$=0.1, so we only use $\beta$=0.3 in our experimentAnd $p_{k,j}$ means proportion of the instances of class k to client j. According to the above policy division data, our local data in each client is Non-IID distributed, and in some classes, each client can have relatively few or no data samples. As mentioned above, we set the number of clients to 10 by default.

In addition, we also use three networks with different architectures as the base encoder:\\
(1)CNN network: It has two 5x5 convolution layers followed by 2x2 max pooling (the first with 6 channels and the second with 16 channels) and two fully connected layers with ReLU activation (the first with 120 units and the second with 84 units).\\
(2)ResNet18\cite{b14}: See TABLE.~\ref{tab:resnetarch} for specific network structure.\\
(3)ResNet50\cite{b14}: See TABLE.~\ref{tab:resnetarch} for specific network structure.

\newcommand{\blocka}[2]{\multirow{3}{*}{\(\left[\begin{array}{c}\text{3$\times$3, #1}\\[-.1em] \text{3$\times$3, #1} \end{array}\right]\)$\times$#2}
}
\newcommand{\blockb}[3]{\multirow{3}{*}{\(\left[\begin{array}{c}\text{1$\times$1, #2}\\[-.1em] \text{3$\times$3, #2}\\[-.1em] \text{1$\times$1, #1}\end{array}\right]\)$\times$#3}
}
\renewcommand\arraystretch{1.1}
\setlength{\tabcolsep}{2pt}
\begin{table}[h]
\begin{center}
\resizebox{1\linewidth}{!}{
\begin{tabular}{c|c|c|c}
\hline
layer name & output size & 18-layer & 50-layer \\
\hline
conv1 & 112$\times$112 & \multicolumn{2}{c}{7$\times$7, 64, stride 2}\\
\hline
\multirow{4}{*}{conv2\_x} & \multirow{4}{*}{56$\times$56} & \multicolumn{2}{c}{3$\times$3 max pool, stride 2} \\\cline{3-4}
  &  &   \blocka{64}{2}  & \blockb{256}{64}{3}\\
  \cr  \\
\hline
\multirow{3}{*}{conv3\_x} & \multirow{3}{*}{28$\times$28}  & \blocka{128}{2}  & \blockb{512}{128}{4} 
  \cr  \cr \\
\hline
\multirow{3}{*}{conv4\_x} & \multirow{3}{*}{14$\times$14}  & \blocka{256}{2} & \blockb{1024}{256}{6} \\
  \cr \\
\hline
\multirow{3}{*}{conv5\_x} & \multirow{3}{*}{7$\times$7}  & \blocka{512}{2} & \blockb{2048}{512}{3}\\
  \cr \\
\hline
& 1$\times$1  & \multicolumn{2}{c}{average pool, 1000-d fc, softmax} \\
\hline
\multicolumn{2}{c|}{FLOPs} & 1.8$\times10^9$  & 3.8$\times10^9$  \\
\hline
\end{tabular}
}
\end{center}
\vspace{-.5em}
\caption{Architectures for ImageNet. Down-sampling is performed by conv3\_1, conv4\_1, and conv5\_1 with a stride of 2.}
\label{tab:resnetarch}
\vspace{-.5em}
\end{table}

For all datasets, like \cite{b15}, we use a 2-layer MLP as the projection head. The output dimension of the projection head is set to 256 by default. Note that all baselines use the same network architecture as FedPDC
(including the projection head) for fair comparison.

We used SGD with a learning rate of 0.01 for all methods. The SGD weight attenuation is set to 0.00001 and the SGD momentum is set to 0.9. The batch size is set to 64. The local epochs for solo is set to 300. For all federated learning methods, the number of local epochs is set to 10, unless explicitly specified.

\subsection{Accuracy Comparison}\label{d.3}

For FedPDC, we tune hyperparameter $\mu$ from \{0.1, 1, 5, 10\} and find the best result is $\mu=10$. We think this phenomenon occurs because our local model tends to converge in the middle and late training period, so we use a larger $\mu$ to correct the update direction, so we get better results. In MOON and FedProx, there are also hyperparameter that can be adjusted manually. We have also made a series of adjustments to achieve high accuracy. For FedProx, hyperparameter $\mu$  controls the
weight of its proximal term (i.e., $L_{FedProx}=\ell_{FedAvg}+\mu\ell_{FedProx}$). We tune µ from {0.001, 0.01, 0.1, 1}(the range is also used in the previous paper \cite{b16}) and report the best result. For MOON, hyperparameter $\mu$ controls the penalty weight of the similarity between the local model and the global model(i.e., $L_{MOON}=\ell_{CrossEntropyLoss}+\mu\ell_{MOON}$). We tune $\mu$ from \{0.1, 1, 5, 10\} and report the best result.

\begin{table}[]
\centering
\caption{The top1 accuracy of FedPDC and the other baselines on CIFAR10, we use Dirichlet distribution $\beta=\{0.1, 0.3\}$  For FedPDC, MOON, FedAvg and FedProx. For SOLO, we report the mean among all parties.}
\label{tbl:cifar10acc}
\resizebox{\linewidth}{!}{
\begin{tabular}{|c|c|c|c|}
\hline
Method & CNN, b=0.1 & CNN, b=0.3 & ResNet50, b=0.1   \\ \hline \hline
\textbf{FedPDC} & \textbf{64.3\%} & \textbf{67.4\%} & \textbf{86.1\%}    \\ \hline
MOON & 64.0\% & 66.8\% & 83.2\%    \\ \hline
FedAvg & 62.1\% & 66.1\% & 84.0\%    \\ \hline
FedProx & 62.9\% & 66.3\% & 84.2\%  \\ \hline
SOLO &28.4\% &\textbackslash{}  &\textbackslash{}    \\\hline
\end{tabular}
}
\end{table}

\begin{table}[]
\centering
\caption{The top-1 accuracy of FedPDC and the other baselines on CIFAR-100, we use Dirichlet distribution $\beta=\{0.1, 0.3, 0.5\}$  For FedPDC, MOON, FedAvg and FedProx. For SOLO, we report the mean among all parties.}
\label{tbl:cifar100acc}
\resizebox{\linewidth}{!}{
\begin{tabular}{|c|c|c|c|}
\hline
Method & CNN, b=0.1 & CNN, b=0.3 & CNN, b=0.5   \\ \hline \hline
\textbf{FedPDC} & \textbf{28.8\%} & \textbf{31.3\%} & \textbf{31.5\%}    \\ \hline
MOON & 28.8\% & 30.7\% & 31.4\%    \\ \hline
FedAvg & 28.3\% & 30.3\% & 31.2\%     \\ \hline
FedProx & 28.0\% & 29.5\% & 31.2\%   \\ \hline
SOLO &15.8\% &\textbackslash{} &\textbackslash{}   \\\hline
\end{tabular}
}
\end{table}

\begin{table}[]
\centering
\caption{The top-1 accuracy of FedPDC and the other baselines on Tiny-Imagenet, we use Dirichlet distribution $\beta=0.3$  For FedPDC, MOON, FedAvg and FedProx.}
\label{tbl:Tiny-Imagenetacc}
\resizebox{\linewidth}{!}{%
\begin{tabular}{|c|c|c|c|c|}
\hline
Method & \textbf{FedPDC} & MOON & FedAvg & FedProx \\ \hline \hline
resNet18, b=0.3 & \textbf{24.03\%} & 23.67\% & 23.35\% & 17.58\% \\ \hline
\end{tabular}
}
\end{table}

Using the above settings, we obtained the test accuracy of each algorithm. See TABLE.~\ref{tbl:cifar10acc} for details. As a result, our algorithm FedPDC maintains almost the highest accuracy compared to the rest of the algorithms when the data is Non-IID. And we can see that the more unbalanced the data distribution, the stronger the advantage of our algorithm. The accuracy of SOLO is much lower than that of other federated learning algorithms, which shows that all federated learning algorithms can effectively improve the model accuracy of the client. FedPDC offers a significant improvement in accuracy over FedAvg in all tasks. The overall accuracy of FedProx is similar to that of FedAvg, and even lower sometimes. This shows that the penalty term adopted by FedProx has little improvement in accuracy in actual experiments. Moreover, the convergence speed of FedProx is obviously slower than that of other algorithms, and the adjustment of hyperparameter $\mu$ has a great impact on the accuracy.

\subsection{Communication Efficiency}\label{d.4}

Previous studies have shown that the main factors affecting the federated learning speed are the communication time and the size of the transmission model\cite{b5}. Therefore, compared with other algorithms, FedPDC does not add a large number of parameters to the transmission model, so the communication efficiency is the same as the above three algorithms.

In the figure below, we show the test accuracy curves of each round of the above three algorithms on each data set. Because FedPDC is not sensitive to the hyperparameter $\mu$, we can see that its convergence rate is similar to FedAvg, especially on CIFAR-10 and CIFAR-100. In most cases, FedPDC converges faster than FedAvg due to the improvement of the aggregation method. But then FedPDC can get the corrected aggregation and data balance comparison loss from the server, so its subsequent convergence accuracy will be significantly higher than FedAvg.

\begin{figure}[ht]
    \centering
    \includegraphics[scale=0.3]{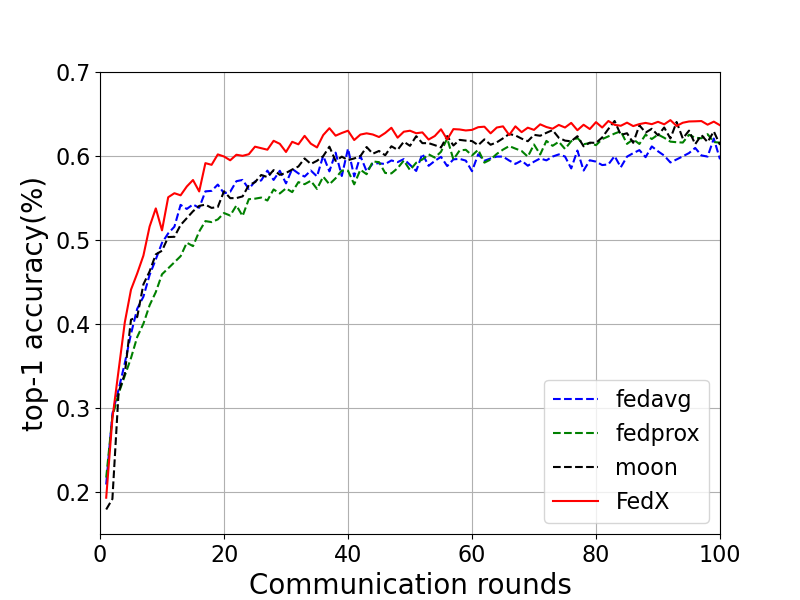}
    \caption{Test accuracy and communication round table of various algorithms in CIFAR-10 dataset.}\label{fig3}
\end{figure}
\begin{figure}[ht]
    \centering
    \includegraphics[scale=0.3]{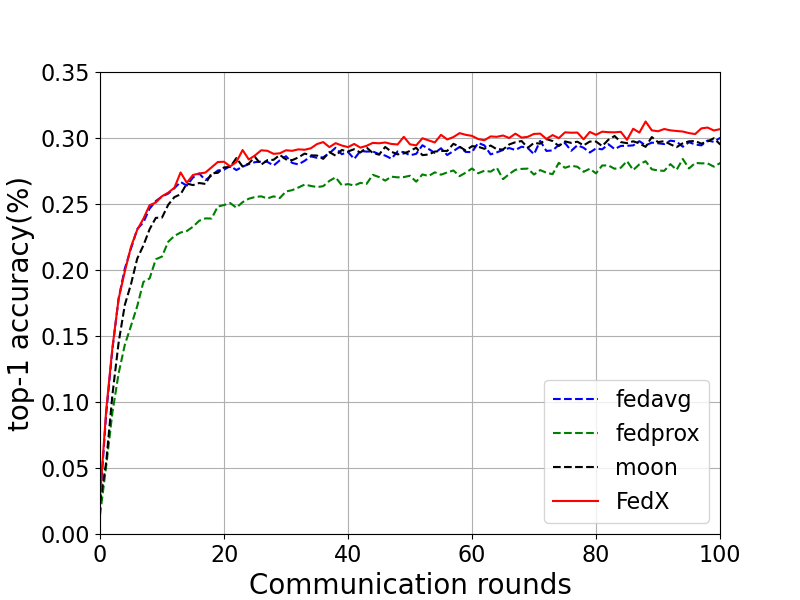}
    \caption{Test accuracy and communication round table of various algorithms in CIFAR-100 dataset.}\label{fig4}
\end{figure}
\begin{figure}[ht]
    \centering
    \includegraphics[scale=0.3]{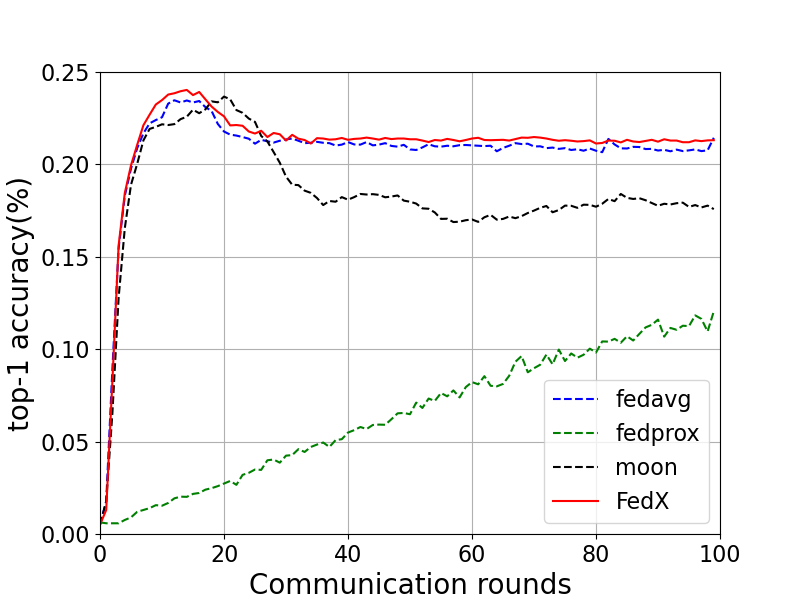}
    \caption{Test accuracy and communication round table of various algorithms in Tiny-Imagenet dataset.}\label{fig4}
\end{figure}
\begin{table}[]
\centering
\caption{The number of rounds of different approaches to achieve the same accuracy as running FedAvg for 100 rounds (CIFAR-10/100) or 20 rounds (Tiny-Imagenet). The speedup of an approach is computed against FedAvg.}
\label{tbl:comm}
\resizebox{\linewidth}{!}{%
\begin{tabular}{|c|c|c|c|c|c|c|}
\hline
 \multirow{2}{*}{Method} & \multicolumn{2}{c|}{CIFAR-10} & \multicolumn{2}{c|}{CIFAR-100} & \multicolumn{2}{c|}{Tiny-Imagenet} \\ \cline{2-7} 
 & \#rounds & speedup & \#rounds & speedup & \#rounds & speedup \\ \hline \hline
FedAvg & 100 & 1$\times$ & 100 & 1$\times$ &20  & 1$\times$ \\ \hline
FedProx & 56 & 1.8$\times$ & 95 & 1.1$\times$ & \textbackslash{}  & \textless{}1$\times$ \\ \hline
MOON & 37 & 2.7$\times$ & 78 & 1.3$\times$ & 19 & 1.1$\times$ \\ \hline
FedPDC & 25 & 4$\times$ & 56 & 1.8$\times$  & 10 & 2$\times$  \\ \hline
\end{tabular}
}
\end{table}

\section{Further work}\label{f}
\begin{algorithm}[h]  
  \caption{\large self-adaption-FedPDC}  
  \label{alg:Framwork}  
  \begin{algorithmic}  
    \Require  
    number of communication rounds T, number of parties N, number of local epochs E, temperature $\tau$ ,learning rate $\eta$ , hyper-parameter $\mu$  ,server-train dataset S(s)
    \Ensure  
  The final model $w^{T}$ \\
  \textbf{server executes:}
      \For{t = 0, 1, ..., T - 1}
      \State test $w_{i}^{t}$ in S(s) $\longrightarrow$ local accuracy $p_{i}^{t}$
      \For{i = 1, 2, ..., N \textbf{in parallel}} 
      \State send the global model $w^{t}$ , global accuracy $p_{t}$ to $N_{i}$ 
      \State $w_{i}^{t}$ ← \textbf{PartyLocalTraining}(i, $w^{t}$, $p_{t}$)
      \EndFor
      \State $w^{t+1} \leftarrow \sum_{k=1}^{N} \frac{\left|p_{t}^{i}\right|}{\left|\sum_{i=1}^{\mathcal{N}} p_{t}^{i}\right|} w_{k}^{t}$
      \EndFor
      \State \textbf{return $w^{T}$}
      \State \textbf{PartyLocalTraining(i, $w^{t}$, $p_{t}$):}
      \State $w_{i}^{t} \leftarrow w^{t}$
      \For{epoch i = 0, 1, ..., E}
      \For{each batch \textbf{b} = \{x, y\} of $D^{i}$}
      \State $\ell_{sup} \leftarrow \text { CrossEntropyLoss }\left(F_{w_{i}^{t}}(x), y\right)$
      \State $\ell_{con} \leftarrow p_{t}$
      \State $\ell \leftarrow \ell_{\text {sup }}+0.5\times t \ell_{\text {con }}$
      \State $w_{i}^{t} \leftarrow w_{i}^{t}-\eta \nabla \ell$
      \EndFor
      \EndFor
      \State \textbf{return $w_{i}^{t}$ to server}
  \end{algorithmic}  
\end{algorithm}

In the above, our algorithm calculates the loss function $\ell_{con}$ in local training. A super parameter $\lambda$ is involved. In order to improve the usability and robustness of our algorithm, we further propose an adaptive hyperparameter method. Generally, the test accuracy of our model will be greatly improved in the previous rounds of training. Therefore, in the early stage of training, we prefer the local model to learn as much as possible from the characteristics of the local data set, rather than rushing to do too much correction. In the middle of the training, the prediction accuracy of our model began to slow down with the increase of the number of rounds. At this time, we hope that the local model will be corrected more strongly than at the beginning of training, so the hyperparameter will become larger accordingly. At the end of the training, the model tends to converge, and the change range of prediction accuracy is very small. Therefore, we hope that the local model can be corrected most, that is, the hyperparameter is the largest in the whole training.

Because we found that the size of the hyperparameter is negatively related to the number of training rounds, we can simply make an adaptation according to the total number of training rounds:

\begin{equation}
 		\label{eq9}
\lambda = 0.5\times n
\end{equation}

Indicates that this round is the $n$th communication round. Therefore, the above algorithm can be further expressed as:

\section{Conclusion}\label{g}

Federated learning has become a hot research area with privacy as an important consideration, and has been widely used in medical, financial, and image recognition fields. Although there are many previous algorithms, such as MOON, FedProx and so on, to solve the non-independent and identical distribution of data in each client, the performance of FedPDC is significantly improved when the data distribution in the client is extremely unbalanced. FedPDC introduces a new federated learning concept, which verifies the local model on the server's independent identically distributed data set, and aggregates and adds penalty items accordingly. Our extensive experiments have proved that FedPDC has greatly improved in image recognition tasks compared with other algorithms, and because FedPDC does not need to input images, it can also be used for various non image tasks. Finally, we also propose a simple adaptive scheme for the hyperparameters of FedPDC, which can get a simpler parameter adjustment process and a better model effect. At the same time, we also prove the convergence of FedPDC, and prove that its convergence has theoretical basis.

\onecolumn

\appendix
\section{Complete proof}
\setcounter{equation}{0}
\setcounter{subsection}{0}
\renewcommand{\theequation}{A.\arabic{equation}}
\renewcommand{\thesubsection}{A.\arabic{subsection}}
\subsection{Proof of Theorem 1}
First, by definition, we have
\begin{equation}
 		\label{eqA.1}
\nabla F_{k}\left(w_{k}^{t+1}\right)+\mu\left(w_{k}^{t+1}-w^{t}\right)=0
\end{equation}
Then we can define $\bar{w}^{t+1}=\mathbb{E}_{k}\left[w_{k}^{t+1}\right]$, based on this definition, we can conclude that
\begin{equation}
 		\label{eqA.2}
\bar{w}^{t+1}-w^{t}=\frac{-1}{\mu} \mathbb{E}_{k}\left[\nabla F_{k}\left(w_{k}^{t+1}\right)\right]
\end{equation}
We further define $\bar{\mu}=\mu-L_{-}>0$, $\bar{\mu}\le\mu$, so we can get
\begin{equation}
 		\label{eqA.3}
\left\|\bar{w}_{k}^{t+1}-w^{t}\right\| \leqslant\frac{1}{\bar{\mu}}\left\|\nabla F_{k}\left(w^{t}\right)\right\|
\end{equation}
Therefore,
\begin{equation}
 		\label{eqA.4}
\left\|\bar{w}^{t+1}-w^{t}\right\| \leqslant \mathbb{E} _{k}\left[\left\|w_{k}^{t+1}-w^{t}\right\|\right] \leqslant \frac{1}{\bar{\mu}} \mathbb{E} _{k}\left[\left\|\nabla F_{k}\left(w^{t}\right)\right\|\right] \leqslant \frac{1}{\bar{\mu}} \sqrt{\mathbb{E} _{k}\left[\left\|\nabla F_{k}(w^{t})\right\| ^{2}\right]} 
\end{equation}
According to the boundedness assumption of $B$, the above equation can be further deduced as
\begin{equation}
 		\label{eqA.5}
\left\|\bar{w}^{t+1}-w^{t}\right\| \leqslant \frac{B}{\bar{\mu}}\left\|\nabla f\left(w^{t}\right)\right\|
\end{equation}
Now we define $M_{t+1}$, such that $\bar{w}^{t+1}-w^{t}=\frac{-1}{\mu}\left(\nabla f\left(w^{t}\right)+M_{t+1}\right)$, so $M_{t+1}=\mathbb{E}_{k}\left[\nabla F_{k}\left(w_{k}^{t+1}\right)-\nabla F_{k}\left(w^{t}\right)\right]$. We can bound $\left\|M_{t+1}\right\|$:
\begin{equation}
 		\label{eqA.6}
\left\|M_{t+1}\right\| \leqslant\mathbb{E} _{k}\left[L\left\|w_{k}^{t+1}-w_{k}^{t}\right\|\right] \leqslant\frac{L}{\bar{\mu}} \times \mathbb{E} _{k}\left[\left\|\nabla F_{k}\left(w^{t}\right)\right\|\right] \leqslant\frac{L}{\bar{\mu}} B\left\|\nabla f\left(w^{t}\right)\right\|
\end{equation}
where the last inequality is also due to the bounded dissimilarity assumption.
Based on the L-Lipschitz smoothness of $f$ and Taylor expansion, we have
\begin{equation}
 		\label{eqA.7}
\begin{aligned}
f\left(\bar{w}^{t+1}\right) & \leqslant f\left(w^{t}\right)+\left\langle\nabla f\left(w^{t}\right), \bar{w}^{t+1}-w^{t}\right\rangle+\frac{L}{2}\left\|\bar{w}^{t+1}-w^{t}\right\|^{2}  \\
& \leqslant f\left(w^{t}\right)-\frac{1}{\mu}\left\|\nabla f\left(w^{t}\right)\right\|^{2}-\frac{1}{\mu}\left\langle\nabla f\left(w^{t}\right), M_{t+1}\right\rangle+\frac{L B^{2}}{2 \bar{\mu}^{2}}\left\|\nabla f\left(w^{t}\right)\right\|^{2} \\
& \leqslant f\left(w^{t}\right)-\left(\frac{1}{\mu}-\frac{L B}{\bar{\mu} \mu}-\frac{L B^{2}}{2 \bar{\mu}^{2}}\right) \left\|\nabla f\left(w^{t}\right)\right\|^{2} 
\end{aligned}
\end{equation}
It can be seen from the above inequality that if we set the penalty parameter $\mu$ large enough, we can get the decrease of the target value of $f\left(\bar{w}^{t+1}\right)-f\left(w^{t}\right)$, which is proportional to $\left\|\nabla f\left(w^{t}\right)\right\|^{2}$. But this is not the way FedPDC works. In this algorithm, only some clients are selected to approximate $\bar{w}^{t}$. So we use the Lipschitz continuity of function $f$ to find $\mathbb{E}\left[f\left(w^{t+1}\right)\right]$. 
\begin{equation}
 		\label{eqA.8}
f\left(w^{t+1}\right) \leq f\left(\bar{w}^{t+1}\right)+L_{0}\left\|w^{t+1}-\bar{w}^{t+1}\right\|
\end{equation}
where $L_{0}$ is the Lipschitz continuous constant of the local function $f$, then we have
\begin{equation}
 		\label{eqA.9}
\begin{aligned}
L_{0} & \leq\left\|\nabla f\left(w^{t}\right)\right\|+L \max \left(\left\|\bar{w}^{t+1}-w^{t}\right\|,\left\|w^{t+1}-w^{t}\right\|\right) \\
& \leq\left\|\nabla f\left(w^{t}\right)\right\|+L\left(\left\|\bar{w}^{t+1}-w^{t}\right\|+\left\|w^{t+1}-w^{t}\right\|\right)
\end{aligned}
\end{equation}
Therefore, if we take expectation with respect to the choice of devices in round $t$ we need to bound
\begin{equation}
 		\label{eqA.10}
\mathbb{E}_{S_{t}}\left[f\left(w^{t+1}\right)\right] \leq f\left(\bar{w}^{t+1}\right)+Q_{t}
\end{equation}
where $Q_{t}=\mathbb{E}_{S_{t}}\left[L_{0}\left\|w^{t+1}-\bar{w}^{t+1}\right\|\right]$. The expectation is the expectation of randomly selected clients. 
\begin{equation}
 		\label{eqA.11}
\begin{aligned}
Q_{t} & \leq \mathbb{E}_{S_{t}}\left[\left(\left\|\nabla f\left(w^{t}\right)\right\|+L\left(\left\|\bar{w}^{t+1}-w^{t}\right\|+\left\|w^{t+1}-w^{t}\right\|\right)\right) \times\left\|w^{t+1}-\bar{w}^{t+1}\right\|\right] \\
& \leq\left(\left\|\nabla f\left(w^{t}\right)\right\|+L\left\|\bar{w}^{t+1}-w^{t}\right\|\right) \mathbb{E}_{S_{t}}\left[\left\|w^{t+1}-\bar{w}^{t+1}\right\|\right]+L \mathbb{E}_{S_{t}}\left[\left\|w^{t+1}-w^{t}\right\| \cdot\left\|w^{t+1}-\bar{w}^{t+1}\right\|\right] \\
& \leq\left(\left\|\nabla f\left(w^{t}\right)\right\|+2 L\left\|\bar{w}^{t+1}-w^{t}\right\|\right) \mathbb{E}_{S_{t}}\left[\left\|w^{t+1}-\bar{w}^{t+1}\right\|\right]+L \mathbb{E}_{S_{t}}\left[\left\|w^{t+1}-\bar{w}^{t+1}\right\|^{2}\right]
\end{aligned}
\end{equation}
Then
\begin{equation}
 		\label{eqA.12}
\mathbb{E}_{S_{t}}\left[\left\|w^{t+1}-\bar{w}^{t+1}\right\|\right] \leq \sqrt{\mathbb{E}_{S_{t}}\left[\left\|w^{t+1}-\bar{w}^{t+1}\right\|^{2}\right]}
\end{equation}
And from (A.5) we can further deduce
\begin{equation}
 		\label{eqA.13}
\begin{aligned}
\mathbb{E} _{s t}\left[\left\|w^{t+1}-\bar{w}^{t+1}\right\|^{2}\right] & \leqslant \frac{1}{k} \mathbb{E} _{k}\left[\left\|w_{k}^{t+1}-{\bar{w}^{t+1}}\right\|^{2}\right] \\
& \leqslant \frac{2}{k}\left[\mathbb{E}  _ { k } [ \| w _ { k } ^ { t + 1 } - w ^ { t } \| ^ { 2 } \right] \\
& \leq \frac{2}{k} \times \frac{1}{\bar{\mu}^{2}}\mathbb{E} _{k}\left[\left\|\nabla F_{k}\left(w^{t}\right)\right\|^{2}\right] \\
& \leqslant \frac{2 B^{2}}{k \bar{\mu}^{2}}\|\nabla f(w ^ { t })\|^{2} 
\end{aligned}
\end{equation}
The first inequality is to randomly select $K$ clients to get the result of $w ^ { t }$, and the last inequality is due to the bounded dissimilarity assumption. Take it into equation (A.11) to get
\begin{equation}
 		\label{eqA.14}
Q_{t} \leqslant\left(\frac{2 L B^{2}}{k \bar{\mu}^{2}}+\left(1+\frac{2 L B}{\bar{\mu}}\right) \frac{\sqrt{2} B}{\bar{\mu} \sqrt{k}}\right)\|\nabla f(w^{t})\|^{2}
\end{equation}
So we finally get
\begin{equation}
 		\label{eqA.14}
\mathbb{E} _{s t}\left[f\left(w^{t+1}\right)\right] \leqslant f\left(w^{t}\right)-\left[\frac{1}{\mu}-\frac{L B}{\bar{\mu} \mu}-\frac{L B^{2}}{2 \bar{\mu}^{2}}-\frac{2 L B^{2}}{k\bar{\mu}^{2}}+\left(1+\frac{2 L B}{\bar{\mu}}\right) \frac{\sqrt{2} B}{\bar{\mu} \sqrt{k}}\right]\left\|\nabla f\left(w^{t}\right)\right\|^{2}
\end{equation}

\begin{thebibliography}{00}
\setlength{\itemsep}{2mm}
\normalsize
\bibitem{b1} Mcmahan H B ,  Moore E ,  Ramage D , et al. Communication-Efficient Learning of Deep Networks from Decentralized Data[C]// 2016.
\bibitem{b2} Li T ,  Sahu A K ,  Zaheer M , et al. Federated Optimization in Heterogeneous Networks[J].  2018.
\bibitem{b3} Li Q ,  He B ,  Song D . Model-Contrastive Federated Learning[J].  2021.
\bibitem{b4} Yang Q ,  Liu Y ,  Chen T , et al. Federated Machine Learning: Concept and Applications[J]. ACM Transactions on Intelligent Systems and Technology, 2019, 10(2):1-19.
\bibitem{b5} Zhu L ,  Lin H ,  Lu Y , et al. Delayed Gradient Averaging: Tolerate the Communication Latency for Federated Learning[C]// Neural Information Processing Systems. 2021.
\bibitem{b6} Karimireddy S P ,  Kale S ,  Mohri M , et al. SCAFFOLD: Stochastic Controlled Averaging for On-Device Federated Learning[J].  2019.
\bibitem{b7} Reisizadeh A ,  Mokhtari A ,  Hassani H , et al. FedPAQ: A Communication-Efficient Federated Learning Method with Periodic Averaging and Quantization[C]// arXiv. arXiv, 2019.
\bibitem{b8} Ling, Zhiwei et al. “FedEntropy: Efficient Device Grouping for Federated Learning Using Maximum Entropy Judgment.” ArXiv abs/2205.12038 (2022): n. pag.
\bibitem{b9} Duan M ,  Liu D ,  Chen X , et al. Self-balancing Federated Learning with Global Imbalanced Data in Mobile Systems[J]. IEEE Transactions on Parallel and Distributed Systems, 2020, PP(99):1-1.
\bibitem{b10} Advances and Open Problems in Federated Learning[J].  2019.
\bibitem{b11} Deng J ,  Dong W ,  Socher R , et al. ImageNet: a Large-Scale Hierarchical Image Database[C]// 2009 IEEE Computer Society Conference on Computer Vision and Pattern Recognition (CVPR 2009), 20-25 June 2009, Miami, Florida, USA. IEEE, 2009.
\bibitem{b12} Yurochkin M ,  Agarwal M ,  Ghosh S , et al. Bayesian Nonparametric Federated Learning of Neural Networks[J].  2019.
\bibitem{b13} Wang H ,  Yurochkin M ,  Sun Y , et al. Federated Learning with Matched Averaging[C]// International Conference on Learning Representations. 2020.
\bibitem{b14} He K, Zhang X, Ren S, et al. Deep residual learning for image recognition[C]//Proceedings of the IEEE conference on computer vision and pattern recognition. 2016: 770-778.
\bibitem{b15} Chen T ,  Kornblith S ,  Norouzi M , et al. A Simple Framework for Contrastive Learning of Visual Representations[J].  2020.
\bibitem{b16} Li T ,  Sahu A K ,  Zaheer M , et al. Federated Optimization in Heterogeneous Networks[J].  2018.
\bibitem{b17} Li X, Huang K, Yang W, et al. On the convergence of fedavg on Non-IID data[J]. arXiv preprint arXiv:1907.02189, 2019.
\bibitem{b18} Ganta D P, Gupta H D, Sheng V S. Knowledge Distillation via Weighted Ensemble of Teaching Assistants[C]//2021 IEEE International Conference on Big Knowledge (ICBK). IEEE, 2021: 30-37.
\bibitem{b19}  Li, D., Wang, J.: Fedmd: Heterogenous federated learning via model distillation. arXiv preprint arXiv:1910.03581 (2019)
\bibitem{b20} Model compression, Cristian Buciluˇa, Rich A Caruana and Alexandru Niculescu-Mizil (2006).
\bibitem{b21} Distilling the Knowledge in a Neural Network, Geoffrey Hinton, Oriol Vinyals and Jeff Dean (2015).
\bibitem{b22} M. G. Arivazhagan, V. Aggarwal, A. K. Singh, and S. Choudhary. Federated learning with personalization layers. arXiv, 2019.
\bibitem{b23} L. Collins, H. Hassani, A. Mokhtari, and S. Shakkottai. Exploiting shared representations for personalized federated learning. In ICML, pages 2089–2099, 2021.
\bibitem{b24} T. Li, S. Hu, A. Beirami, and V. Smith. Ditto: Fair and robust federated learning through personalization. In ICML, pages 6357–6368, 2021.
\bibitem{b25} P. P. Liang, T. Liu, L. Ziyin, R. Salakhutdinov, and L.-P. Morency. Think locally, act globally: Federated learning with local and global representations. arXiv, 2020.
\bibitem{b26} Y. Mansour, M. Mohri, J. Ro, and A. T. Suresh. Three approaches for personalization with applications to federated learning. arXiv, 2020.
\bibitem{b27} K. Matsuda, Y. Sasaki, C. Xiao, and M. Onizuka. Fedme: Federated learning via model exchange. SDM, pages 459–467, 2022.
\bibitem{b28} T. Shen, J. Zhang, X. Jia, F. Zhang, G. Huang, P. Zhou, K. Kuang, F. Wu, and C. Wu. Federated mutual learning. arXiv, 2020.
\bibitem{b29} M. Zhang, K. Sapra, S. Fidler, S. Yeung, and J. M. Alvarez. Personalized federated learning with first order model optimization. In ICLR, 2021.
\bibitem{b30} Rauniyar, Ashish, et al. "Federated Learning for Medical Applications: A Taxonomy, Current Trends, and Research Challenges." arXiv e-prints (2022): arXiv-2208.
\end{thebibliography}
\end{document}